# CLIMB: Controllable Longitudinal Brain Image Generation using Mamba-based Latent Diffusion Model and Gaussian-aligned Autoencoder

**Duy-Phuong Dao[1], Muhammad Taqiyuddin[1], Jahae Kim[1,2] Sang-Heon Lee[3], Hye-Won Jung[4], Jaehoo Choi[5], and Hyung-Jeong Yang[1,*]**

[1] Department of Artificial Intelligence Convergence, Chonnam National University, South Korea

[2] Department of Nuclear Medicine, Chonnam National University, South Korea

[3] University of South Australia, Australia

[4] AdelaideMRI, Australia

[5] Electronics and Telecommunications Research Institute, South Korea

**\* Correspondence: hjyang@jnu.ac.kr**

**Abstract:** Latent diffusion models (LDM) have emerged as powerful generative models in medical imaging, enabling the synthesis of high-quality brain magnetic resonance imaging (MRI) scans. In particular, predicting the evolution of a patient's brain can aid in early intervention, prognosis, and treatment planning. In this study, we introduce CLIMB, Controllable Longitudinal brain Image generation via MamBa-based latent diffusion model, an advanced framework for modeling temporal changes in brain structure. CLIMB is designed to model the structural evolution of the brain structure over time, utilizing a baseline MRI scan and its acquisition age as foundational inputs. Additionally, multiple conditional variables, including projected age, gender, disease status, genetic information, and brain structure volumes, are incorporated to enhance the temporal modeling of anatomical changes. Unlike existing LDM methods that rely on self-attention modules, which effectively capture contextual information from input images but are computationally expensive, our approach leverages Mamba, a state space model architecture that substantially reduces computational overhead while preserving high-quality image synthesis. Furthermore, we introduce a Gaussian-aligned autoencoder (GATE) that extracts latent representations conforming to prior distributions without the sampling noise inherent in conventional variational autoencoders. We train and evaluate our proposed model on the Alzheimer's Disease Neuroimaging Initiative (ADNI) dataset, consisting of 6,306 MRI scans from 1,390 participants. By comparing generated images with real MRI scans, CLIMB achieves a structural similarity index (SSIM) of 0.9433, demonstrating notable improvements over existing state-of-the-art methods. Code is found in Github.

## 1. Introduction

Longitudinal medical image generation, aiming to synthesize follow-up images with a baseline image to simulate anatomical changes over time, has been increasingly explored in medical imaging research [1], [2]. By generating realistic follow-up images, researchers can handle missing data, improve disease progression modeling, and support clinical decision-making [3], [4]. This approach is particularly valuable in neurodegenerative diseases (e.g., Alzheimer's), where understanding temporal anatomical changes is crucial for early diagnosis and treatment planning.

Over the past decade, the emergence of generative artificial intelligence (AI) models has paved the way for the generation of realistic and high-quality images across various fields [5]–[8], especially in medical imaging [1], [2], [9], [10]. Diffusion models (DM) have set a new benchmark in generative AI, surpassing traditional generative adversarial networks (GANs) [11]–[13] and variational autoencoders (VAEs) [14] in both image quality and diversity. However, conventional diffusion models suffer from high computational costs and slow inference due to their iterative denoising process in pixel space, making them resource-intensive and inefficient. Latent diffusion models (LDMs) offer a solution to these challenges by conducting operations in a lower-dimensional latent space, which substantially decreases computational requirements while preserving high-quality image generation capabilities [1], [6].

Most existing LDM approaches [1], [6], [10] employ VAE [15] to compress input data into a lower-dimensional latent space where the diffusion model is subsequently applied. VAEs facilitate the learning of a structured latent space conforming to a Gaussian distribution. However, during VAE training, latent features are sampled from a prior distribution, which introduces stochastic noise that often results in reconstructions lacking sharpness or clarity. To address this issue, sliced Wasserstein distance [16] was proposed to align the latent space distribution more closely with the prior distribution (e.g., Gaussian) without incorporating sampling noise, thereby preserving the fidelity and sharpness of the reconstructed images.

Furthermore, current LDMs [1], [17] architectures predominantly rely on trans- former-based backbones [18], which excel at capturing contextual relationships within input data. However, this comes with a significant computational cost due to the extensive dimensionality of the latent space, particularly when processing three-dimensional medical images. Additionally, transformer-based models typically require extensive training data to achieve optimal performance. Recently, state space models (SSMs) [19], [20], particularly Mamba [21], have been designed to be more computationally efficient than transformers while still maintaining competitive performance. Notably, Mamba can work well with small datasets, offering a more efficient alternative. While global context is crucial in medical image analysis, local features are equally significant for processing medical imaging data. Integrating both aspects can enhance model performance considerably.

Besides, recent studies [1], [10] have shown that incorporating multiple factors—such as age, gender, and brain structure volumes—improves the synthesis of MRI scans. Furthermore, baseline MRI scans serve as a reference point, enabling models to capture personalized anatomical structures and pathological changes in the brain over time. In this study, we introduce CLIMB, a controllable longitudinal brain image generation using Mamba-based latent diffusion model, to synthesize the anatomical structure changes using multiple conditional factors. Our contributions can be summarized as follows:

- We propose CLIMB model that synthesize follow-up MRI scans using a baseline MRI scan with its acquisition age and projected data. It should be noted that we utilize our previously developed IRLSTM model [3] to predict the projected data at follow-up time points.
- We implement sliced Wasserstein distance to align the latent features during the training of our autoencoder model, preserving image fidelity.
- We use the Mamba architecture [21] as the backbone of our latent diffusion model to effectively capture global context while reducing computational complexity compared to the transformer mechanism.

The remainder of this paper is structured as follows: Section II reviews related works in the field. Section III details the architecture of our proposed model. Section IV describes the dataset and presents the experiments. Visualization and discussion are provided in Section V. Finally, Section VI concludes our proposed model.

## 2. Related Work

**Conventional Diffusion Models:** Denoising diffusion probabilistic models (DDPMs) have recently become a cornerstone of generative modeling, offering improved diversity, stability, and fidelity over VAEs and GANs. In medical imaging, diffusion models have been adapted for a variety of synthesis tasks, including MRI synthesis, super-resolution, denoising, and cross-modal translation. These works demonstrate the ability of diffusion to capture complex anatomical distributions and maintain realistic fine-scale textures—an essential property for clinical interpretability. Early efforts employ full-resolution diffusion on 2D slices or small 3D patches to mitigate computational overhead. For instance, several studies apply 2.5D conditioning to generate T1-weighted MRIs from FLAIR or CT modalities, while others leverage noise-conditioned score networks for MRI super-resolution and PET-to-MRI translation. Despite these successes, conventional diffusion models remain computationally expensive for high-dimensional 3D medical volumes, as both training and inference require iterative denoising across thousands of steps. Additionally, directly applying pixel-space diffusion limits scalability and temporal modeling, making it impractical for longitudinal forecasting tasks that require repeated sampling over time.

**Latent Diffusion:** To address the memory and runtime bottlenecks of full-resolution diffusion, latent diffusion models (LDMs) compress input images into a lower-dimensional latent space before applying the diffusion process. This strategy preserves semantic and structural fidelity through a perceptually aligned autoencoder while dramatically reducing computational cost. Recent works demonstrate that latent diffusion achieves comparable or superior image realism with 10–20 times lower GPU memory consumption than conventional diffusion, enabling the generation of full 3D MRI volumes and multi-contrast studies. Some approaches integrate spatial attention and anatomical priors in the latent domain to improve fine detail reconstruction. However, applications of latent diffusion to temporal or longitudinal settings remain limited. Most medical LDMs focus on static synthesis or modality translation, without mechanisms to model temporal evolution, enforce cross-time consistency, or condition on clinical variables such as age, genetics, sex, or brain region volumes. Extending LDMs to longitudinal forecasting thus presents an open opportunity to leverage their efficiency and controllability for dynamic, multi-timepoint imaging tasks. Moreover, in most implementations, this autoencoder follows the variational autoencoder (VAE) framework, which has become the standard choice for LDMs due to its stability, smooth latent manifold, and probabilistic formulation. The VAE objective encourages the latent distribution to approximate a Gaussian prior, enabling efficient sampling and ensuring that the latent space is continuous-properties that make it well-suited for diffusion training. However, this design also introduces **inherent limitations**. The stochastic sampling of latent features during VAE training—where latent codes are drawn from a Gaussian distribution rather than deterministically encoded—can produce **overly smooth reconstructions**, especially when the latent dimensionality is heavily compressed.

**Image-to-Image Forecasting:** aims to predict future medical images given baseline observations, effectively modeling disease progression or anatomical change over time. SADM [2] integrates transformer-based attention module into DPM as conditional embedding to generate a sequence of MRI scans. Recently inspired by [10], [26], [1] propose BrLP that use multiple conditional covariables such as age, gender, disease status, and brain structure volumes to synthesize MRI scans over time. BrLP also incorporates baseline MRI scans and the patient's age at the time of capture to control structural changes over time. However, even though BrLP use LDM to reduce model complexity and inference time, their model also consumes a significant memory due to the use of self-attention (SA) to capture global context of input that the latent features are still large. To address this limitation, emerging studies explore state-space models (SSMs) as an alternative to attention for modeling long-range dependencies. Unlike SA, which requires global pairwise interactions, SSMs parameterize sequence dynamics through recurrent hidden states, enabling linear-time complexity and improved scalability for high-dimensional or temporal data. However, existing SSM-based approaches have yet to be fully integrated into longitudinal medical image forecasting, where balancing global spatial context and efficient temporal modeling remains an open challenge.

Our work bridges the gap between diffusion-based generative modeling and longitudinal forecasting. We integrate latent diffusion with a regularized autoencoder to construct a compact, high-fidelity latent space that mitigates the blurring effects of conventional VAE encoders. Incorporating state-space modeling improves temporal scalability beyond self-attention. Combined with explicit clinical conditioning, our framework enables controllable and probabilistic MRI forecasting that preserves anatomical consistency while maintaining computational efficiency across extended temporal horizons.

## 3. Proposed Methods

In this section, we describe the overall architecture of our method, as illustrated in Figure 1. We provide a detailed description of our proposed CLIMB model for synthesizing anatomical changes over time. CLIMB comprises two main training stages: an autoencoder and a latent diffusion process. Initially, we briefly describe how diffusion and Mamba mechanisms work. We then introduce our proposed GATE model, which extracts and aligns the latent space to follow a Gaussian distribution. Next, we present the CLIMB architecture and explain how it integrates a baseline MRI scan, its acquisition age, and projected data as conditional factors. Finally, we describe the inference process to generate a future MRI scan.

### 3.1 Preliminary

**Latent Diffusion Models (LDM)** involve compressing high-dimensional data into a lower-dimensional latent space using a VAE before applying the diffusion process within this latent space. The diffusion model consists of two main processes: the forward process and the reverse process. Given input x, pre-trained encoder E, and decoder D, the latent space $z_0$ = E(x) is gradually added noise over T steps until it converges to a noisy latent variable $z_T$ ~ N(0, 1): $q(z_t|z_0)$ = N($z_t$; $\sqrt{\bar{\alpha}_t}z_0$, $1 - \bar{\alpha}_t$), where $\bar{\alpha}_t$ is a constant at each step t. The $z_t$ can be sampled as $z_t = \sqrt{\bar{\alpha}_t}z_0 + \sqrt{1 - \bar{\alpha}_t}\epsilon_t$ where $\epsilon_t$ ~ N(0, 1). The reverse process is then trained to denoise the added noise from the forward process at each step t via a trainable network θ($z_t$, t): $p_\theta(z_{t-1}|z_t)$ = N($z_{t-1}$; $\mu_\theta(z_t, t)$, $\sum_\theta(z_t, t)$). Finally, the denoised latent space is fed into the decoder D to generate new data. The objective function of the diffusion process can be defined as follows [5]:

$$\mathcal{L} = ||\varepsilon_t - \varepsilon_\theta(\sqrt{\bar{\alpha}_t}z_0 + \sqrt{1 - \bar{\alpha}_t}\varepsilon_t)||^2 \qquad (1)$$

### 3.2 Gaussian-aligned Autoencoder

To enable a diffusion model to operate effectively in the latent domain, the latent representation must be both continuous and statistically aligned with a Gaussian prior. Diffusion models assume that data are gradually transformed into Gaussian noise in the forward process and recovered from it in the reverse process. Therefore, a latent space that already conforms to a Gaussian structure significantly simplifies the learning dynamics and ensures compatibility between the autoencoding and diffusion stages.

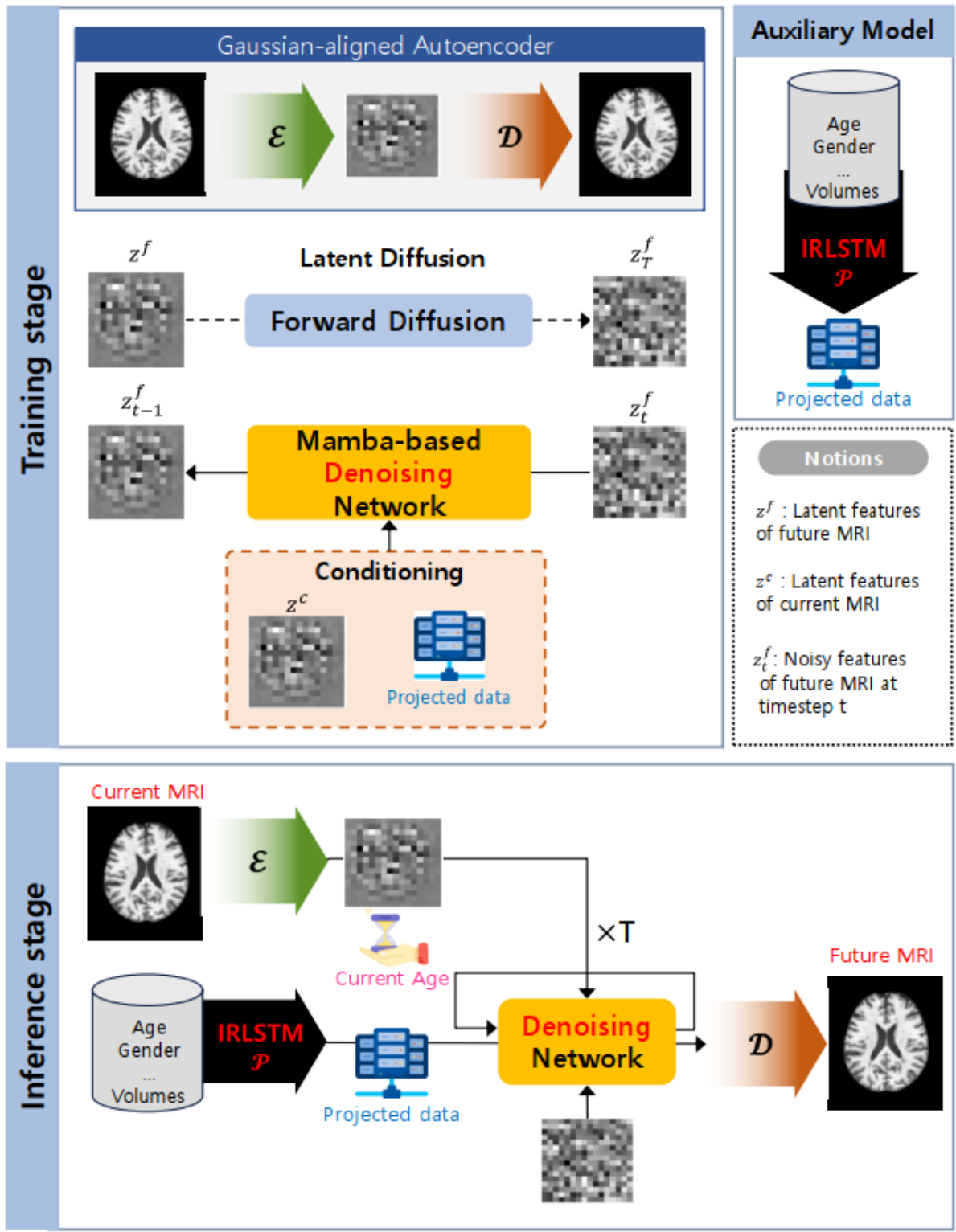

**Figure 1.** The overall architecture of our method. E and D are the encoder and decoder of the autoencoder model, respectively. P denotes the IRLSTM model [3] which is used to predict disease status and brain structure volumes at the projected age.

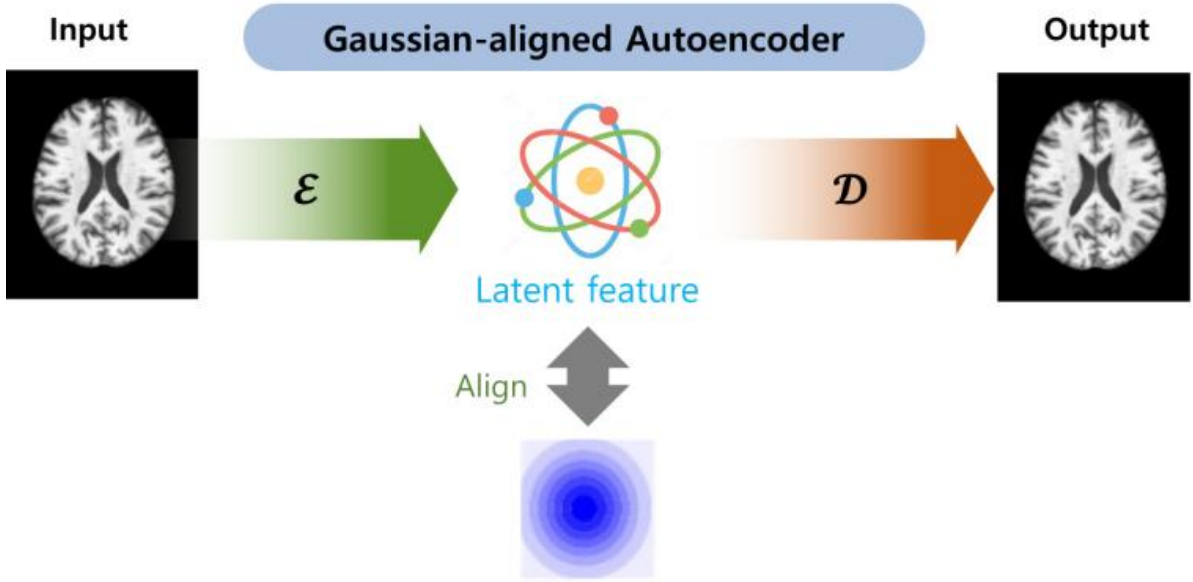

**Figure 2.** The overall architecture of the Gaussian-aligned autoencoder. The encoder E and decoder D are based on the ResNet architecture..

In our framework, we introduce a Gaussian-Aligned Autoencoder (GATE) that enforces Gaussian structure on the latent representation without stochastic sampling or KL-based variational regularization. Unlike Variational Autoencoders, which inject random noise into the latent variables via the reparameterization trick, the proposed approach adopts a deterministic distribution matching strategy based on sliced cumulative density function (CDF) alignment.

Let $x \in X$ denote the input sample, and $z = E(x) \in Z$ the encoded latent feature, where $E(\cdot)$ and $D(\cdot)$ denote the encoder and decoder networks. The reconstruction is defined as $\hat{x} = D(z)$, and the reconstruction loss is given by:

$$\mathcal{L}_{\text{rec}} = \mathbb{E}_{x \sim p_{\text{data}}(x)} \left[ \|x - \hat{x}\|_2^2 \right].$$

To regularize the latent distribution $q(z)$ toward a standard Gaussian prior $p(z) = \mathcal{N}(0, I)$, we use a discrete sliced CDF alignment procedure. Specifically, given the latent dimensionality $d$, we randomly sample $K$ (K << d) projection directions $\{v_k\}_{k=1}^{K}$ uniformly distributed on the unit hypersphere $\mathbb{S}^{d-1}$. Each projection direction is obtained by sampling a random vector $w \sim N(0, I_d)$ and normalizing it by its L2 norm:

$$\mathbf{v}_k = \frac{\mathbf{w}_k}{\|\mathbf{w}_k\|_2}, \quad \text{for } k = 1, \ldots, K.$$

For each projection direction $v_k$, we compute one-dimensional projected samples for both the encoded latent variables and the Gaussian prior:

$$s_i^{(k)} = \langle z_i, \mathbf{v}_k \rangle, \quad r_i^{(k)} = \langle \tilde{z}_i, \mathbf{v}_k \rangle,$$

Where $z_i \sim q(z)$, $\tilde{z}_i \sim p(z)$ and $s_{(i)}^{(k)}$ and $r_{(i)}^{(k)}$ denote the ordered statistics (sorted values) of the two sets of samples along direction $v_k$.

Finnally, we compute distance between CDF's latent feature and prior Gaussian as follows:

$$\mathcal{L}_{\mathrm{SD}} = \frac{1}{K} \sum_{k=1}^{K} \mathbb{E}\left[ \left| s_{(i)}^{(k)} - r_{(i)}^{(k)} \right|^p \right].$$

To further enhance the quality and realism of reconstructions, we incorporate perceptual and adversarial objectives in the training of the decoder. The perceptual loss encourages high-level feature similarity between the reconstructed and original images, measured in the feature space of a pretrained visual network (e.g., VGG or CLIP):

$$\mathcal{L}_{\mathrm{perc}} = \mathbb{E}_{x \sim p_{\mathrm{data}}(x)} \left[ \| \phi(x) - \phi(\hat{x}) \|_2^2 \right],$$

where $\phi(\cdot)$ denotes the feature extraction function from an intermediate layer of a pretrained model. This loss ensures that reconstructions maintain semantic and structural consistency, even when pixel-level differences remain.

In addition, we employ an adversarial loss to promote photo-realistic outputs. A discriminator network $D_{\mathrm{adv}}$ is trained alongside the autoencoder to distinguish between real samples $x$ and reconstructed samples $\hat{x}$. The generator (autoencoder decoder) aims to fool the discriminator, yielding the adversarial objective:

$$\mathcal{L}_{\mathrm{adv}} = \mathbb{E}_{x \sim p_{\mathrm{data}}(x)} \left[ \log D_{\mathrm{adv}}(x) \right] + \mathbb{E}_{x \sim p_{\mathrm{data}}(x)} \left[ \log(1 - D_{\mathrm{adv}}(\hat{x})) \right].$$

Combining these components, the total training loss of the GATE model becomes:

$$\mathcal{L}_{\mathrm{GATE}} = \lambda_{\mathrm{rec}} \mathcal{L}_{\mathrm{rec}} + \lambda_{\mathrm{SD}} \mathcal{L}_{\mathrm{SD}} + \lambda_{\mathrm{perc}} \mathcal{L}_{\mathrm{perc}} + \lambda_{\mathrm{adv}} \mathcal{L}_{\mathrm{adv}}$$

where each $\lambda$ −term controls the relative contribution of its respective loss.

This multi-objective design ensures that the autoencoder not only learns a Gaussian-aligned latent space compatible with diffusion processes but also generates high-quality, perceptually faithful, and realistic reconstructions. The combination of sliced density alignment, perceptual consistency, and adversarial realism provides a robust foundation for latent diffusion training.

*3.3 State space-based Denoising Network*

In this section, we propose a state space-driven conditional diffusion network that integrates image-conditioned and temporal information to model complex latent dynamics efficiently. The overall architecture follows a U-Net–style encoder–decoder design, where the SSM is adopted as a lightweight yet expressive alternative to traditional SA mechansim. The proposed framework takes as input a noisy latent representation $z_t' \in \mathbb{R}^{C \times D \times H \times W}$, generated at timestep $t$ of the diffusion process, along with auxiliary projected data and image features representing conditioning attributes (e.g., current age). The network progressively denoises $z_t'$ to produce $z_{t-1}'$, representing a less noisy latent approximation. As shown in Fig. 3, the architecture is composed of an encoder path, a mid block, and a decoder path. The encoder compresses spatial–temporal information through successive down-sampling blocks, while the decoder reconstructs the latent representation using up-sampling and skip connections. Conditioning information from the projected data and image features is injected through Cross-Attention operations, allowing the model to align its generation process with external cues.

At the core of our model lies the state space mechanism, which replaces traditional Self-Attention within each diffusion block. Unlike attention-based models that rely on quadratic complexity to capture token-to-token interactions, Mamba leverages the expressive formulation of state-space models (SSMs) to achieve efficient, context-aware computation. Conventional SSMs model one-dimensional sequential data by transforming input sequences $x_t \in R^M$ into output sequences $y_t \in R$ through a hidden state $h_t \in R^N$, governed by matrices $A$, $B$, and $C$:

$$B = Linear_B(x), \quad C = Linear_C(x)$$

$$\Delta = Softplus(Linear_\Delta(x))$$

$$\bar{A} = exp(\Delta A), \quad \bar{B} = (\Delta A)^{-1}(exp(\Delta A) - I) \cdot \Delta B \quad (3)$$

$$h_t = \bar{A}h_{t-1} + \bar{B}x_t, \quad y_t = Ch_t \quad (4)$$

From Equation 4, the output sequence $y_t$ can be computed using a convolution operation, allowing for parallel execution to reduce running time. The kernel $\bar{K}$ of the convolution is formulated as follows [21]:

$$\bar{K} = (C\bar{B}, C\bar{A}\bar{B}, ..., C\bar{A}^l\bar{B}, ...), \quad y = x * \bar{K} \quad (5)$$

Each term in $\bar{K}$ corresponds to how information from earlier time steps (or spatial positions) influences the current output, effectively encoding long-range dependencies. The convolutional form means this operation can be computed in linear time and parallelized, unlike Self-Attention.

To ensure stable and efficient training, we employ a two-stage learning strategy inspired by [1], [26]. In the first stage, the diffusion model (shown on the left side of Fig. 3) is trained using only the projected variables, enabling it to learn the intrinsic relationships between these conditioning signals and the target latent outputs. Once convergence is achieved, the model's parameters are frozen and duplicated to initialize the image feature processing network (highlighted in blue in Fig. 3). In the second stage, the full model is trained with both inputs, but only the parameters of the image feature network are updated. To facilitate smooth adaptation and preserve the pre-trained diffusion representation, zero-initialized convolutional layers are inserted before and after each block of the image feature network. This approach allows gradual information flow from the image-conditioned pathway, stabilizing the optimization and accelerating convergence.

*3.4 Inference Process*

Given a current MRI, we use the encoder of the GATE model to extract image features. In parallel, the IRLSTM model predicts projected data based on variables such as age, gender, current disease status, and current brain volumes. These predicted features are then fed into the Mamba-based denoising network to sample the latent representation of a future MRI scan. The sampled latent features are subsequently passed through the decoder of the GATE model to generate the corresponding future MRI. It is important to note that we adopt the Denoising Diffusion Implicit Model [31] in the reverse process of latent diffusion to enable faster sampling. The number of sampling steps is set to 25. Additionally, we employ latent average sampling by generating 10 samples per run, ensuring robustness and reliability in the final output.

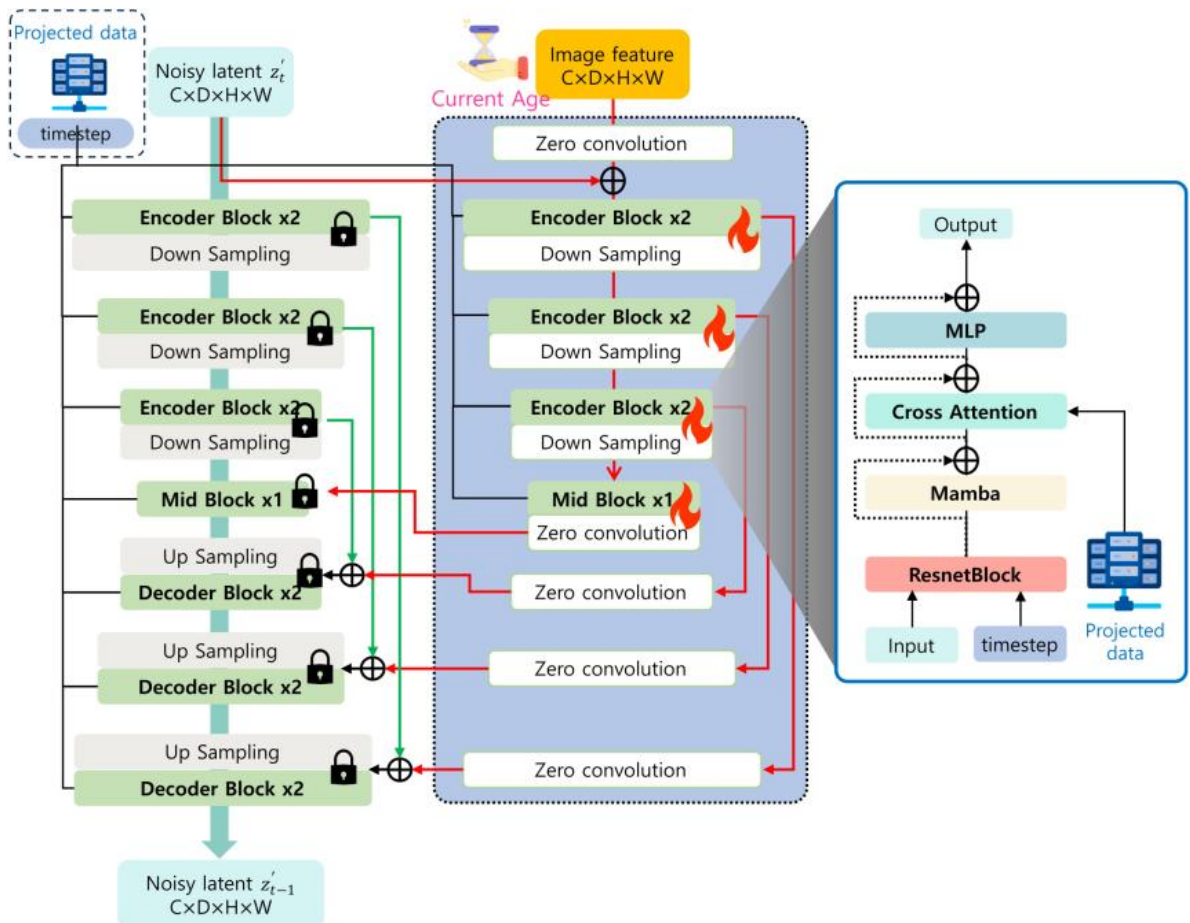

**Figure 3.** The architecture of the Mamba-based denoising network. C, D, H, and W represent the number of channels, depth, height, and width of the image features, respectively.

## 4. Dataset and Experiment Results

### 4.1 Dataset and Experiment Settings

Our research utilized data from the ADNIMERGE dataset, sourced from the Alzheimer's Disease Neuroimaging Initiative (ADNI) database, which consolidates information from four sub-datasets: ADNI-G0, ADNI-1, ADNI-2, and ADNI-3. We selected 1,390 participants across 6306 total visits, comprising 2,095 visits diagnosed as Cognitively Normal (CN), 2,764 as Mild Cognitive Impairment (MCI), 1438 as Alzheimer's Disease (AD), and 9 with a missing diagnosis. Following the methodology outlined in [1], we applied a consistent preprocessing pipeline to all MRI scans, including N4 bias field correction [32], skull stripping [33], affine registration to MNI152 space [34], [35], white matter-based intensity normalization [36], resampling to a voxel size of 1.5mm, and zeros-padding. This results in standardized images with dimension of $128 \times 160 \times 128$. In addition, we used a pretrained SynthSeg model [37] to segment brain regions. From these segmentations, we extracted the volumes of five brain structures highly associated with Alzheimer's disease [38], [39]: the hippocampus, gray matter, white matter, ventricles, and amygdala. The dataset was then divided into three subsets: 80% for training, 5% for validation, and 15% for testing.

*4.2 Comparison with the State-of-the-art Methods*

We performed extensive experiments using an Intel Xeon Gold 6326 CPU and an NVIDIA RTX A6000 GPU to train and evaluate both our proposed model and several baseline methods. All model parameters were optimized using the Adam optimize [40] with a learning rate of 0.0001. For training the GATE model, we used a batch size of 2 and ran the training for 100,000 iterations. Furthermore, for the latent diffusion models, we employed a batch size of 32 and trained the models for 500 epochs.

We assessed the quality of synthesized follow-up MRI scans using four established quantitative metrics: mean square error (MSE), structural similarity (SSIM), peak signal-to-noise ratio (PSNR), and perceptual similarity (LPIPS) [29]. For comparison, we selected the SADM [2] and BrLP [1] models as baselines, as shown in Table I and Figure 4. It is worth noting that we evaluated the pre-trained BrLP model on our test set, which may partially overlap with the training data used in their original study. Additionally, the SADM model originally applies the diffusion technique in voxel space, which led to out-of-memory issues when applied to our dataset. To address this, we followed the BrLP approach and trained the SADM model using a latent diffusion mechanism instead.

As shown in Table I, our CLIMB model clearly demonstrates superior performance across all evaluation metrics compared to existing baseline models. CLIMB significantly reduces MSE (2.01e-3), indicating more accurate pixel-level reconstruction. It also achieves the highest SSIM (0.9433) and PSNR (27.82), reflecting better preservation of image structure and higher overall quality. Notably, our CLIMB method achieves the lowest LPIPS score (0.0587), indicating a significant improvement in perceptual quality compared to all baselines. This strong performance can be attributed to the use of the GATE autoencoder, which preserves finegrained structural and semantic details more effectively than conventional VAE-based latent diffusion models. In terms of model efficiency, our proposed model also runs faster (2.92 seconds per image) and consumes less memory (5699 Mb), indicating its suitability for practical deployment in timesensitive and resource-constrained environments. Based on the results, we can conclude that our CLIMB model has achieved state-of-the-art performance in longitudinal MRI synthesis.

TABLE I

COMPARISON WITH THE BASELINE MODEL AND THE ABLATION STUDY. * INDICATES RESULTS REPORTED IN THE PAPER [1]

| Method | Autoencoder | Diffusion | MSE↓ | SSIM↑ | PSNR↑ | LPIPS↓ | Time (s/image) | Memory (Mb) |
|---|---|---|---|---|---|---|---|---|
| Latent-SADM* [2] | VAE | SA-based | 8.00e-3 | 0.8500 | - | - | - | - |
| Latent-SADM [2] | VAE | SA-based | 4.75e-3 | 0.8818 | 23.70 | 0.1112 | 6.35 | 8879 |
| BrLP* [1] | VAE | SA-based | 4.00e-3 | 0.9100 | - | - | - | - |
| BrLP [1] | VAE | SA-based | 2.97e-3 | 0.9209 | 26.32 | 0.1171 | 4.01 | 9823 |
| CLIMB (Ours) | GATE | Mamba-based | **2.01e−3** | **0.9433** | **27.82** | **0.0587** | **2.92** | **5699** |

# 5. Visualization, Analysis, and Discussion

*5.1 Visualization*

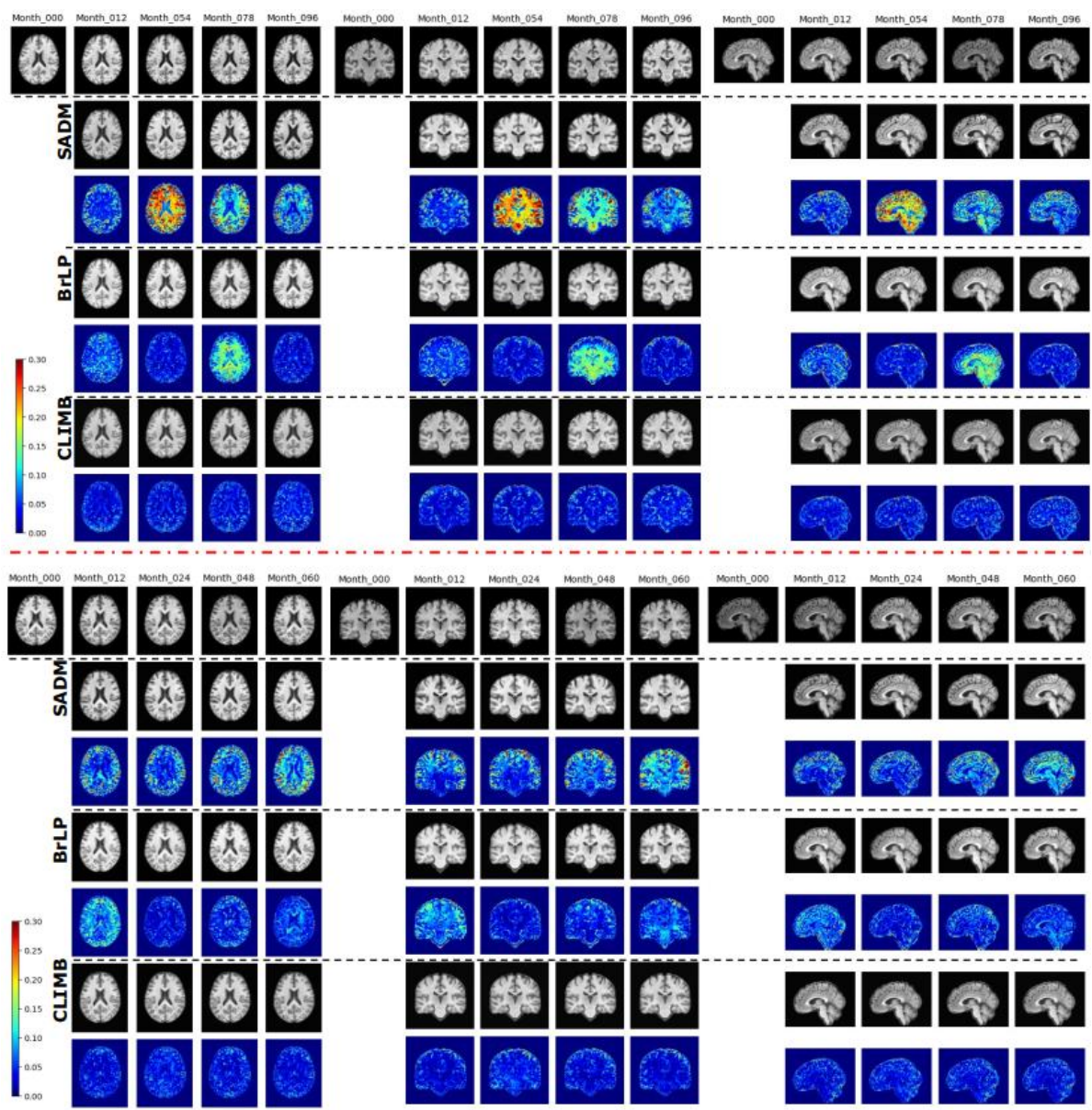

Fig. 4. Samples of generated MRI scans. Jet maps illustrate deviations between real and generated images.

For qualitative comparison, we present several generated MRI samples from both existing models and our proposed CLIMB model in Figure 4. The figure clearly demonstrates that our CLIMB model produces anatomically consistent follow-up MRI scans that closely align with the real MRI scans. Additionally, Figure 4 illustrates that the images generated by CLIMB show less deviation from the actual scans, as visualized in the jet maps (third and fifth rows), compared to those produced by the SADM and BrLP models. These jet maps reveal that SADM often introduces structural artifacts and high error regions, especially in the central brain areas, while BrLP shows moderate improvements but still lacks structural fidelity over long time spans. In contrast, our CLIMB model maintains low reconstruction error across all timepoints and views (axial, coronal, and sagittal), confirming its ability to model realistic longitudinal changes in brain anatomy. For a clearer visualization of how brain structure evolves over time, please refer to our supplementary material or our Github repository.

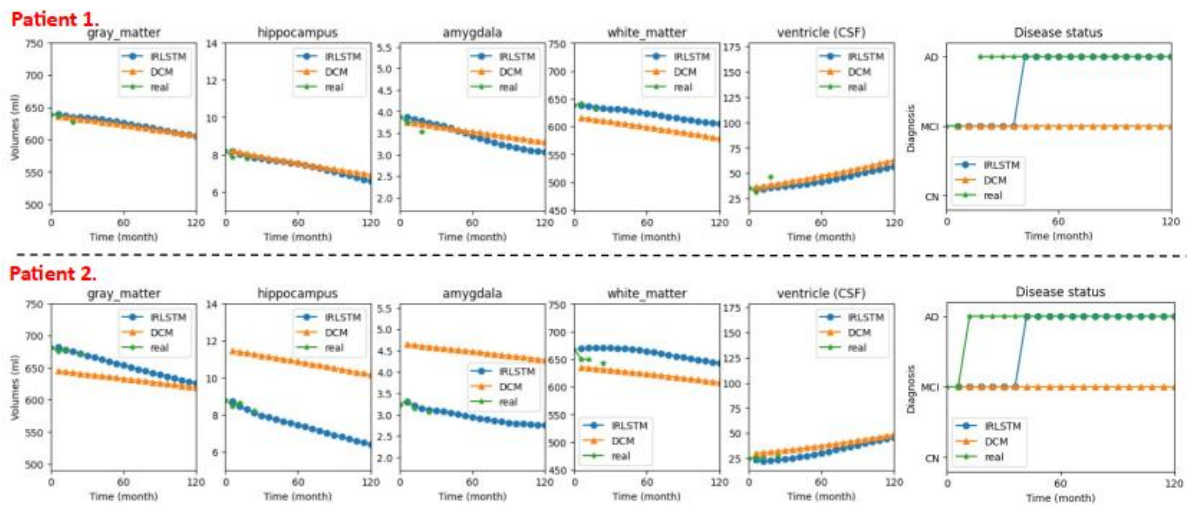

Fig. 5. Several examples of disease status and brain volume in the future.

TABLE II
COMPARISON BETWEEN OUR THE DCM AND IRLSTM METHODS IN TERMS OF PREDICTION OF DISEASE STATUS AND BRAIN VOLUMES. GM, HIP, AMY, WM, AND VEN STAND FOR GRAY MATTER, HIPPOCAMPUS, AMYGDALA, WHITE MATTER, AND VENTRICLE VOLUMES, RESPECTIVELY.

| Method | mAUC | MAE(ml) | | | | |
|---|---|---|---|---|---|---|
| | Diagnosis | GM | Hip | Amy | WM | Ven |
| DCM [1] | 90.13 | 13.12 | 0.30 | 0.28 | 25.28 | 2.83 |
| IRLSTM [3] | 95.38 | 11.39 | 0.28 | 0.15 | 12.18 | 2.72 |

TABLE III
COMPARISON BETWEEN THE VARIATIONAL AUTOENCODER AND OUR GAUSSIAN-ALIGNED AUTOENCODER

| Autoencoder | MSE↓ | SSIM↑ | PSNR↑ | LPIPS↓ |
|---|---|---|---|---|
| VAE | 2.10e-3 | 0.9392 | 27.27 | 0.1080 |
| GATE | 2.01e−3 | 0.9433 | 27.82 | 0.0587 |

In addition, since the predicted disease status and brain volumes are used as conditional factors to synthesize future MRI scans, we conducted additional experiments to compare the performance of disease status and brain volume prediction models, as shown in Table II. For qualitative evaluation, several examples of predicted results from the DCM [1] and IRLSTM [3] models are presented in Figure 5, illustrating predictions of disease status and brain volume in the future. The results demonstrate that the IRLSTM model consistently outperforms the DCM method across all evaluation metrics. This indicates that our previously proposed IRLSTM model is more effective in modeling longitudinal progression and generating reliable predictions for downstream MRI synthesis.

*5.2 Ablation Studies*

Table III demonstrates that our proposed method achieves better performance compared to the variational autoencoder in terms of all evaluation metrics. Particularly, our method achieves a much lower LPIPS score (0.0587) compared to the VAE (0.1080). This is a notable improvement, suggesting that our autoencoder produces reconstructions that are more perceptually similar to the ground truth.

In addition, we conduct experiments to compare the performance of different backbones in the diffusion model, as shown in Table IV. Without using SA or Mamba mechanism , the model (Res+CA) yields the worst results across all metrics, likely due to lacking of global features. Moreover, when replacing SA with the Mamba mechanism (Res+Mamba+CA), the model not only achieves better reconstruction quality but also improves efficiency by reducing both inference time and memory consumption.

TABLE IV
PERFORMANCE COMPARISON OF BACKBONE VARIANTS IN LATENT DIFFUSION MODEL.

| Diffusion | MSE↓ | SSIM↑ | PSNR↑ | LPIPS↓ | Time (s/image) | Memory (Mb) |
|---|---|---|---|---|---|---|
| Res+CA | 2.19e-3 | 0.9417 | 27.37 | 0.0596 | **2.08** | **5547** |
| Res+SA+CA | 2.10e-3 | 0.9418 | 27.62 | 0.0596 | 4.17 | 6477 |
| Res+Mamba+CA | **2.01e−3** | **0.9433** | **27.82** | **0.0587** | 2.92 | 5699 |

TABLE V
COMPARISON OF DIFFERENT HIDDEN STATE SIZES IN OUR LATENT DIFFUSION MODEL.

| Variants | MSE↓ | SSIM↑ | PSNR↑ | LPIPS↓ | Time (s/image) | Memory (Mb) |
|---|---|---|---|---|---|---|
| CLIMB-tiny | 2.14e-3 | 0.9436 | 27.43 | 0.0576 | 2.18 | 4803 |
| CLIMB-base | 2.01e-3 | 0.9433 | 27.82 | 0.0587 | 2.92 | 5699 |
| CLIMB-large | **1.93e−3** | **0.9454** | **27.96** | **0.0569** | 6.22 | 8979 |

Furthermore, we conduct experiments to evaluate the effect of hidden size in our latent diffusion model. As shown in Table V, the CLIMB-large model demonstrates superior performance across all quality metrics, indicating improved image fidelity and perceptual similarity. However, this model also incurs higher inference time and memory consumption. In contrast, the CLIMB-base and CLIMB-tiny models achieve lower performance but are more efficient and consume significantly less memory. These results highlight a trade-off between image quality and computational efficiency.

## 6. Conclusion

In this work, we proposed CLIMB, a longitudinal brain image generation using Mamba-based latent diffusion model, to synthesize the follow-up MRI scans. We proposed a Gaussianaligned autoencoder model to extract the latent space conforming to a Gaussian distribution without sampling any noise. We designed a latent diffusion to capture local and global features from the input data. The experiments demonstrated that our CLIMB model significantly enhance the performance compared with the baseline model across all evaluation metrics. The qualitative findings also indicate that our method can reliably predict longitudinal changes in brain structures, offering potential benefits in reducing patient costs, improving clinical decision-making, and supporting early detection of disease progression.

While SSIM and MSE scores indicate strong performance, these metrics alone are insufficient to validate the analytical efficacy of algorithms in brain image interpretation. Empirical evaluations demonstrate that methods optimized for SSIM/MSE may exhibit diminished accuracy when applied to clinical or anatomical analysis of neuroimaging data, emphasizing the critical need for validation by medical experts, such as radiologists or neurologists.

**Acknowledgments:** This work was supported by the Ministry of Science and ICT (MSIT), Korea, through the Institute of Information & Communications Technology Planning & Evaluation (IITP) under the following programs: Innovative Human Resource Development for Local

Intellectualization (IITP- 2025-RS-2022-00156287), Artificial Intelligence Convergence Innovation Human Resources Development (IITP-2023-RS-2023-00256629), and the Information Technology Research Center (ITRC) support program (IITP-2025-RS-2024-00437718). Additional support was provided by the National Research Foundation of Korea (NRF) grant funded by the Korea government (MSIT) (RS-2023-00208397).

**Conflicts of Interest:**